\documentclass[sigconf]{acmart}

\usepackage{caption}
\usepackage{subcaption}

\AtBeginDocument{%
  \providecommand\BibTeX{{%
    \normalfont B\kern-0.5em{\scshape i\kern-0.25em b}\kern-0.8em\TeX}}}

\copyrightyear{2024}
\acmYear{2024}
\setcopyright{rightsretained}
\acmConference[GECCO '24]{Genetic and Evolutionary Computation Conference}{July 14--18, 2024}{Melbourne, VIC, Australia} \acmBooktitle{Genetic and Evolutionary Computation Conference (GECCO '24), July 14--18, 2024, Melbourne, VIC, Australia} \acmDOI{10.1145/3638529.3654013} \acmISBN{979-8-4007-0494-9/24/07}

\begin{document}

\title{Towards Multi-Morphology Controllers with Diversity and Knowledge Distillation}

\author{Alican Mertan}
\email{alican.mertan@uvm.edu}
\affiliation{%
  \institution{Neurobotics Lab \\ University of Vermont}
  \city{Burlington}
  \state{VT}
  \country{USA}
}

\author{Nick Cheney}
\email{ncheney@uvm.edu}
\affiliation{%
  \institution{Neurobotics Lab \\ University of Vermont}
  \city{Burlington}
  \state{VT}
  \country{USA}}


\begin{abstract}
  Finding controllers that perform well across multiple morphologies is an important milestone for large-scale robotics, in line with recent advances via foundation models in other areas of machine learning. However, the challenges of learning a single controller to control multiple morphologies make the `one robot one task' paradigm dominant in the field. To alleviate these challenges, we present a pipeline that: (1) leverages Quality Diversity algorithms like MAP-Elites to create a dataset of many single-task/single-morphology teacher controllers, then (2) distills those diverse controllers into a single multi-morphology controller that performs well across many different body plans by mimicking the sensory-action patterns of the teacher controllers via supervised learning. The distilled controller scales well with the number of teachers/morphologies and shows emergent properties. It generalizes to unseen morphologies in a zero-shot manner, providing robustness to morphological perturbations and instant damage recovery. Lastly, the distilled controller is also independent of the teacher controllers -- we can distill the teacher’s knowledge into any controller model, making our approach synergistic with architectural improvements and existing training algorithms for teacher controllers.
\end{abstract}


\begin{teaserfigure}
\centering
  \includegraphics[width=1.0\textwidth]{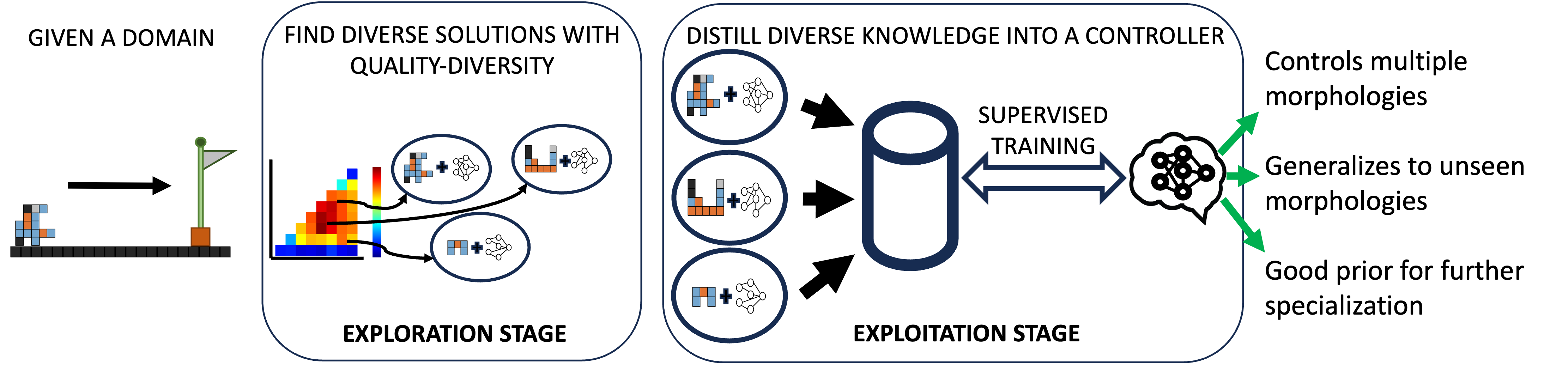}
  \vspace{-0.28in}
  \caption{Proposed pipeline for training a single controller capable of controlling multiple morphologies. Given an environment and a task, Quality Diversity algorithms explore the domain and discover effective morphologies and their controllers. These controllers are used as teacher controllers and their knowledge is distilled into a single controller. The distilled controller successfully controls the morphologies of the teacher controllers. Moreover, the distilled controller generalizes well to unseen morphologies in a zero-shot manner and provides a better prior for further specialization on unseen morphologies and tasks.}
  \Description{Teaser image that shows the proposed pipeline for solving incompatible control.}
  \label{fig:teaser}
  \vspace{1em}
\end{teaserfigure}

\maketitle

\section{Introduction} \label{sect:intro}

Finding controllers that perform well across different morphologies is an important milestone for large-scale robotics. Similar to the `foundation models' that enable progress in other areas of machine learning, such as computer vision or language processing, a foundational multi-morphology controller can facilitate progress in robotics by enabling fine-tuning to downstream tasks with a smaller amount of data (which is important because the best methods for training control models such as evolutionary or reinforcement learning algorithms are data-inefficient). Moreover, models capable of controlling a multitude of robots can enable better brain-body co-optimization by being a good fitness estimator for unseen morphologies~\cite{mertan2024investigating}, hence better specialization and performance can be obtained for a given domain by exploiting multi-morphology controllers. However, obtaining such controllers, despite being of great interest, remains an open problem. As identified in~\cite{gupta2022metamorph}, the field is mostly stuck in the `one robot one task' paradigm -- training a new robot and controller from scratch for each new task such as locomotion, object manipulation, climbing, etc.

The challenges that drive the field into the `one robot one task' paradigm are twofold. The first issue is the incompatible control -- the differences in action and observation spaces of different robots or tasks~\cite{kurin_my_2021}. While it is straightforward to have `compatible' controllers through caching trick~\cite{huang_one_2020}, having wide enough input (and output) layers to accommodate the maximum possible number of sensory inputs (and motor outputs), while simply ignoring unused nodes for morphologies with a smaller number of inputs (outputs), it turns out that such controllers are hard to train to control multiple morphologies, whether reinforcement learning~\cite{huang_one_2020} or evolutionary optimization~\cite{mertan_modular_2023} algorithms are used. This hardship of training stems from the complexity of learning many tasks/morphologies.

These two challenges, incompatible control and hardship of training, are intertwined as the controller model one chooses to use as a multi-morphology controller determines the loss landscape and in return, affects its training dynamics. Previous work on obtaining multi-morphology controllers mostly focuses on complexifying controller models by enforcing modularity (aiming for emergent higher-scale control at the robot level~\cite{nadizar2023fully, huang_one_2020, christensen_distributed_2013, pathak_learning_2019}) or by employing recent developments in Graph Neural Networks and Transformers (utilizing their ability to deal with arbitrary-sized inputs~\cite{kurin_my_2021, gupta2022metamorph, wang2018nervenet}). These attempts not only solve the incompatibility problem but also, presumably, shapes the loss landscape and offer easier training.

In this work, however, we focus solely on the hardship of training multi-morphology controllers, instead of addressing it indirectly by architectural changes, which makes our approach synergistic with existing work. Our proposed method stems from our search for a `simple' procedure for learning a multi-morphology controller. Inspired by existing literature on knowledge distillation~\cite{buciluǎ2006model,hinton2015distilling,rusu2015policy,parisotto2015actor,teh2017distral,medvet2024gp,mace2023quality,faldor2023map} -- learning to match input-output patterns of teacher models via supervised learning, we investigate the use of response-based, offline, multi-teacher policy distillation to learn a single multi-morphology controller. We collect a dataset of $(observation,\allowbreak action)$ pairs from teacher controllers optimized to control single morphology and then employ supervised learning to train a single controller on this collected dataset.

Indeed, we show that knowledge distillation results in controllers that match the performance of teacher controllers on many different morphologies, without needing any complex architecture or message-passing scheme -- just a simple neural network with a single hidden layer that is made compatible with the caching trick can learn to control hundreds of robots for the locomotion task in a matter of minutes on a consumer laptop. We also note that our method is agnostic to the choice of the distilled controller. Once we collect a dataset of $(observation,\allowbreak action)$ pairs, it can be used to train any neural network, including the modular or graph-based models developed to alleviate the challenges of incompatible control. 

It might seem, however, that optimizing many teacher controllers that specialize in particular morphologies to be able to train a single multi-morphology controller defeats the purpose. For this, we would like to show two important points. First, the teacher controllers are only used to collect $(observation, action)$ pairs and then discarded. They can be as simple as possible for faster optimization on a particular morphology. Moreover, one can use abundant publicly available pre-trained controllers instead of training them from scratch. More importantly, we show that the distilled controller displays emergent properties. Our investigations show that the distilled controller is robust to morphological perturbations and generalizes to unseen morphologies in a zero-shot manner. Moreover, it can be used as a prior for further specialization on unseen morphologies or tasks to speed up the adaptation process. These properties justify the cost of training teacher controllers -- we gain more than the sum of teacher controllers by distilling them into a single controller. Most of all, our proposed approach is orthogonal to the previous work developing complex architectures for multi-morphology control -- we could use such models to further enhance the emergent properties of the distilled controllers.

It is not obvious, nonetheless, how to select teacher controllers' morphologies. Indeed, the choice of morphologies to train the multi-morphology controller is an important question that has been ignored in the literature so far. Existing works either heuristically~\cite{christensen_distributed_2013, huang_one_2020, kurin_my_2021, wang2018nervenet} choose the morphologies to train with, or assume that effective morphologies are known prior to the training~\cite{gupta2022metamorph}. Considering that, unlike most prior work, we need experiences ($(observation,\allowbreak action)$ pairs) of effective morphologies as well to distill that knowledge into a single controller, we resort to Quality Diversity (QD) algorithms~\cite{pugh_confronting_2015,pugh_quality_2016} for finding effective morphologies with their optimized controllers, similar to~\cite{mace2023quality,faldor2023map} where QD is used to create a repertoire of behaviors for a single robot and knowledge distillation is used to distill that knowledge into a single controller. In our case, given that the feature descriptors are based on morphological attributes, a QD algorithm is used to explore the morphology space and optimize controllers for a variety of high-performing morphologies with different trade-offs in their morphological attributes for a given domain. We show that these controllers can be used as teachers to distill into a single controller that is capable of controlling a variety of distinct morphologies that perform well for the given domain, and that generalizes well to unseen morphologies in a zero-shot manner. 

Overall, we present the two-stage pipeline illustrated in Fig.~\ref{fig:teaser} that first explores a given domain with QD algorithms to find high-performing morphologies and their respective controllers, and then exploits this knowledge by distilling it into a single controller. The main contributions of this work are to:
\begin{itemize}
    \item demonstrate the effectiveness of knowledge distillation for multi-morphology controller training -- a simple procedure that can train simple models to control multiple morphologies. (Sec.~\ref{sect:kd})
    \item propose the use of QD algorithms for the automated discovery of effective morphologies with their optimized controllers for the distillation process. (Sec.~\ref{sect:qd})
    \item investigate the capabilities of the distillation process and distilled controllers (Sec.~\ref{sect:invest}) and show that 
    \begin{itemize}
        \item the distillation process is controller agnostic, making it synergistic with existing work that develops complex controller models. (Sec.~\ref{subsect:model_independent})
        \item the distilled controllers scale well with the number of teachers. (Sec.~\ref{subsect:scale})
        \item the distilled controllers generalize well to unseen morphologies, justifying the cost of obtaining teacher controllers. (Sec.~\ref{subsect:generalize}) 
        \item the distilled controllers provide a better prior for further specialization, unlocking transfer learning opportunities in the robotics field. (Sec.~\ref{subsect:prior}) 
    \end{itemize}
\end{itemize}

\section{Methods} \label{sect:methods}
\textbf{Simulation.} We use Evolution Gym version 1.0.0 (Evogym)~\cite{bhatia2021evolution} for simulating 2D voxel-based soft bodied robots. The raw observations that the simulation provides are processed to acquire voxel volume ($\in \mathbb{R}$), speed ($\in \mathbb{R}^2$), and material type (one-hot encoded vector $v$ of length 5). We also process the timesteps into a saw wave-shaped time signal by applying $\mod{25}$ to it. While the simulation engine is deterministic, we apply a small noise to the observations sampled from $\mathcal{N}(0,0.01)$ to model sensory noise.

\noindent\textbf{Tasks and performance evaluation.} We mainly experiment in the Walker-v0 environment which consists of a flat surface of length 100 in voxels with a locomotion task where robots try to reach the end of the surface. We also experiment with a similar locomotion task that consists of a soft, dynamic surface called BridgeWalker-v0. In both environments, we use the modified reward function from~\cite{mertan_modular_2023, mertan2024investigating} which encourages the robot to finish the tasks as fast as possible. Due to the stochasticity we injected into the system in terms of observation and action noises, we repeat each fitness evaluation multiple times and take the average. Unless otherwise noted, each simulation is repeated 5 times. 

\noindent\textbf{Robot representation and control.} Robots in Evogym consist of 4 types of materials. There are two materials under active control, one that expands horizontally and one that expands vertically. There are also two passive materials, rigid and elastic, that are under the effects of forces created by active materials and dynamics of the environment. Robots are directly represented as a matrix $R \in T_{H \times W}$ where $T \in$ \{0,1,2,3,4\} encodes the materials (or lack thereof), and we use $(H,W)=(5,5)$, following the practice of limiting the robot design space~\cite{creative_machines_lab_cornell_university_ithaca_ny_evolved_2014, cheney_difficulty_2016, cheney_scalable_2018, kriegman_how_2018, medvet_biodiversity_2021, marzougui_comparative_2022, tanaka_co-evolving_2022,mertan_modular_2023,mertan2024investigating}. Robots are controlled by specifying a scalar action ($\in [0.6,1.6]$) that is used to determine the target length by multiplying the action with the resting length. Controllers are queried every 5\textsuperscript{th} timestep and the last action is repeated for the remaining timesteps to prevent high-frequency dynamics. We apply a small noise to the actions sampled from $\mathcal{N}(0,0.01)$ to model actuator noise.

\noindent\textbf{Controller models.} Throughout the work, we experiment with 3 different controllers -- `Global FC', `Global Tx', and `Modular FC', modeled by different neural network architectures -- `FC' stands for fully connected and `Tx' stands for transformer, and belonging 2 different control paradigms -- global indicates a centralized controller where observations from all voxels are concatenated and consumed at once by the controller to output actions for each voxel, similar to the ones used in~\cite{medvet_evolution_2020, ferigo_evolving_2021, mertan_modular_2023, talamini_evolutionary_2019, talamini_criticality-driven_2021,medvet_impact_2022}, and modular indicates a shared, decentralized controller that observes a local neighborhood and output action for a single voxel, similar to the ones used in~\cite{pigozzi_evolving_2022, medvet_evolution_2020, medvet_biodiversity_2021, huang_one_2020,mertan_modular_2023,mertan2024investigating}. All controllers are made compatible through the use of caching trick and are not conditioned on the morphology explicitly.\footnote{Details of controller architectures can be found in our code repository: \href{https://github.com/mertan-a/towards-multi-morphology-controllers}{mertan-a/towards-multi-morphology-controllers}}

\noindent\textbf{Optimization algorithms.} To show the ineffectiveness of joint training on fixed morphologies, we use the reinforcement learning algorithm Twin Delayed Deep Deterministic policy gradients algorithm (TD3)~\cite{fujimoto2018addressing} (starting from the clean implementation of~\cite{huang2022cleanrl}) and the evolutionary optimization algorithm Age-Fitness Pareto Optimization (AFPO)~\cite{schmidt_age-fitness_2010} (with population size of 16). After demonstrating the ineffectiveness of both approaches, we introduce our approach where we use the Map-Elites algorithm~\cite{mouret_illuminating_2015} as a Quality Diversity algorithm to create an archive of diverse and effective morphologies with their optimized controllers where the two feature descriptors are the number of existing voxels and the number of active voxels. In each generation of the Map-Elites algorithm, we create 16 new solutions from 16 randomly chosen solutions from the map. The offspring are created by mutation only. Following~\cite{bhatia2021evolution,mertan_modular_2023,mertan2024investigating}, robot representations are mutated by changing the material of each voxel with a $10\%$ probability and controllers are mutated by adding a noise sampled from $\mathcal{N}(0,0.1)$. The new solutions are created by mutating either the morphology or the controller, chosen with a 50\% probability, as in~\cite{cheney_scalable_2018,mertan_modular_2023,mertan2024investigating}.

\noindent\textbf{Statistical testing} To measure the statistical significance for distilled controllers, we calculate the relative performance of the distilled controllers compared to teacher controllers and apply one-sample t-test~\cite{student1908probable} with the null hypothesis of the population mean of 1. Where we compare two samples, we use the Wilcoxon Rank Sum test~\cite{wilcoxon1964some}. All comparisons are done with the $p=.05$ threshold.

\section{Knowledge Distillation for Multi-Morphology Control} \label{sect:kd}
\begin{figure}
    \centering
    \begin{subfigure}[t]{0.25\linewidth}
         \centering
         \includegraphics[width=0.6\linewidth]{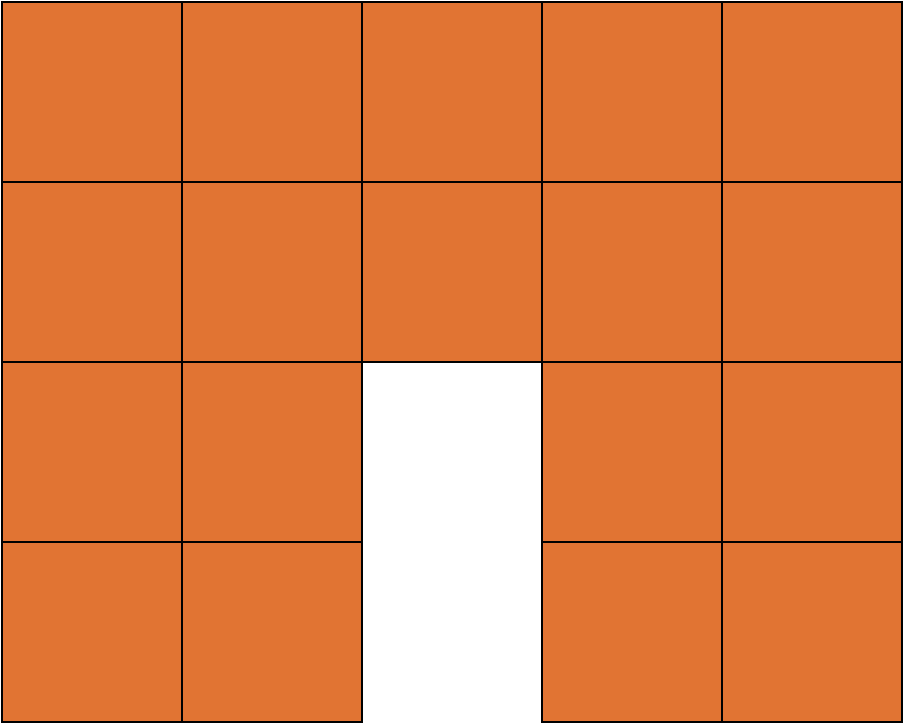}
         \caption{Biped}
         \label{fig:biped}
    \end{subfigure}%
    \begin{subfigure}[t]{0.25\linewidth}
         \centering
         \includegraphics[width=0.6\linewidth]{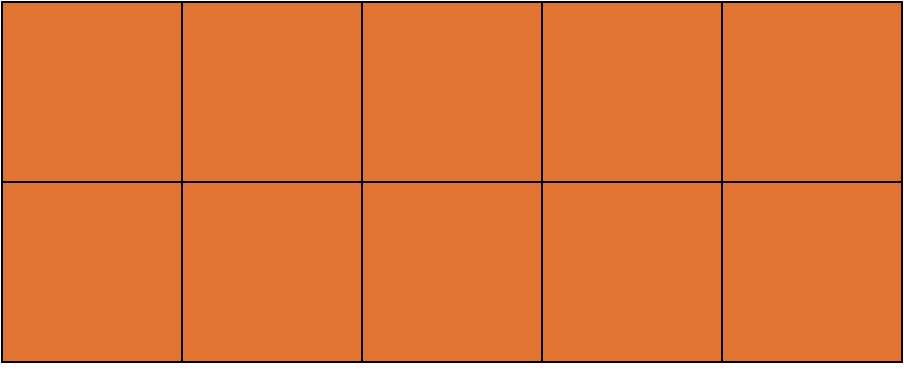}
         \caption{Worm}
         \label{fig:worm}
    \end{subfigure}%
    \begin{subfigure}[t]{0.25\linewidth}
         \centering
         \includegraphics[width=0.6\linewidth]{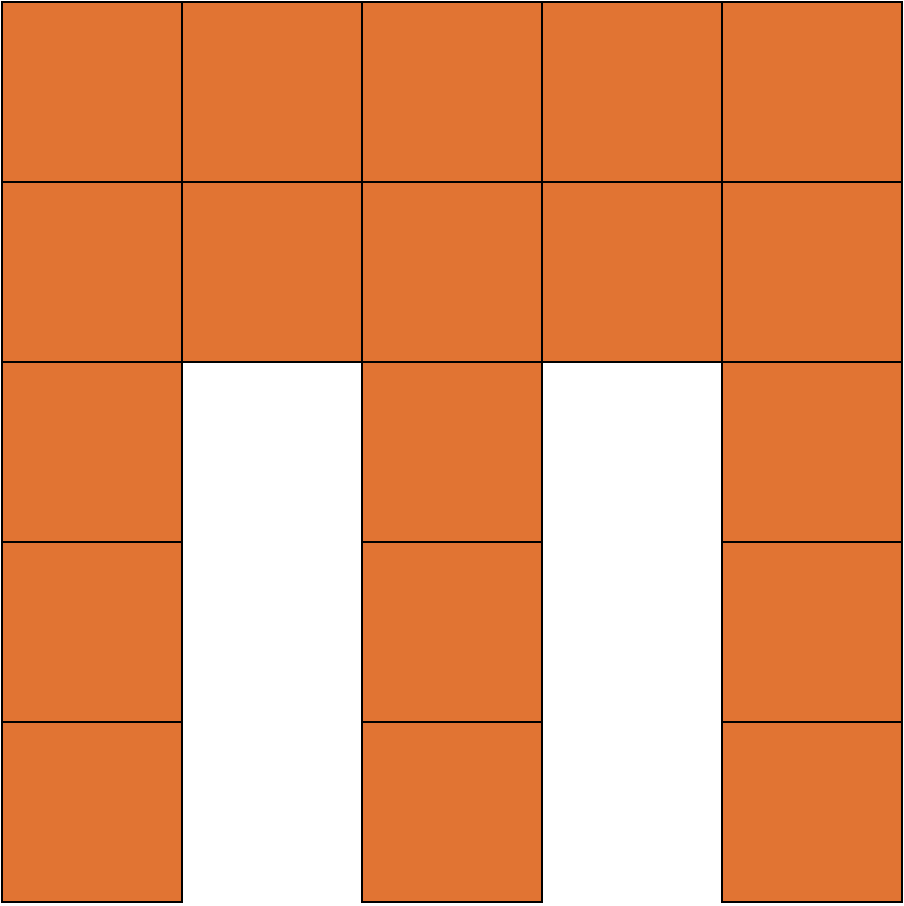}
         \caption{Triped}
         \label{fig:triped}
    \end{subfigure}%
    \begin{subfigure}[t]{0.25\linewidth}
         \centering
         \includegraphics[width=0.6\linewidth]{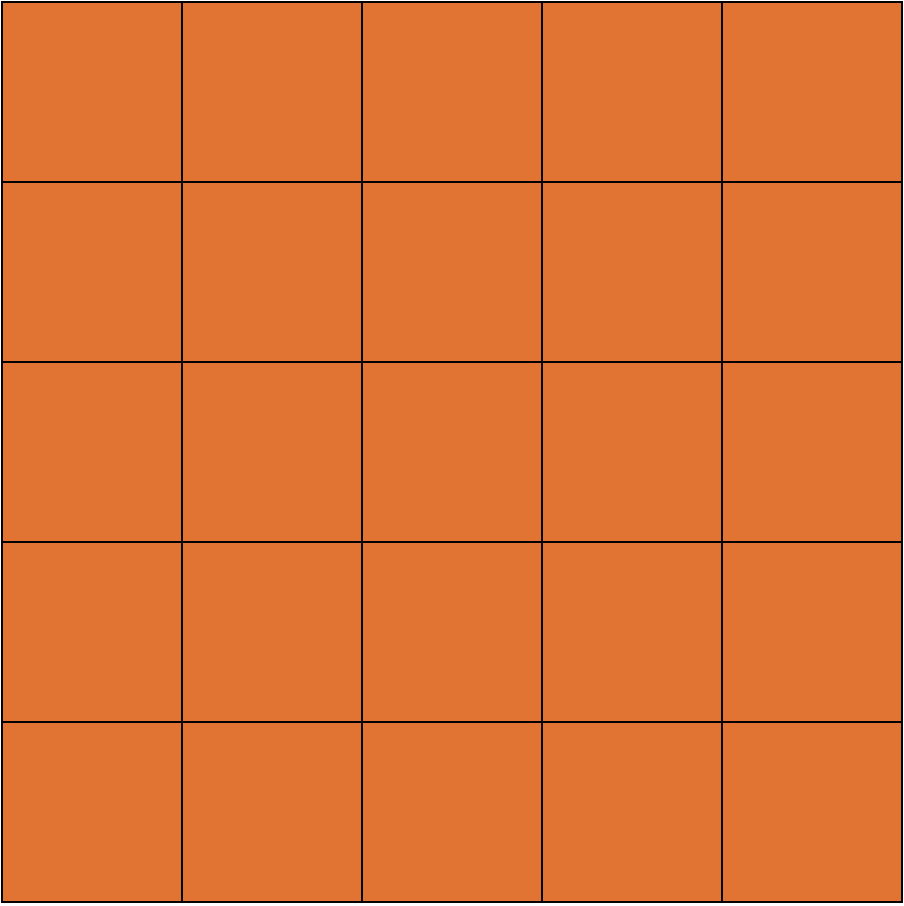}
         \caption{Block}
         \label{fig:block}
    \end{subfigure}
    \caption{Experimented morphologies to show the effectiveness of knowledge distillation for training multi-morphology controller.}
    \label{fig:fixed_ms}
\end{figure}

Optimizing controllers capable of controlling multiple different morphologies is a challenging problem~\cite{huang_one_2020,mertan_modular_2023,gupta2022metamorph}. Here, we provide a naive attempt (referred to as joint training) at learning a controller (Global FC) for controlling the four predetermined and fixed morphologies shown in Fig.~\ref{fig:fixed_ms}, both with evolutionary optimization (AFPO~\cite{schmidt_age-fitness_2010}) and reinforcement learning (TD3~\cite{fujimoto2018addressing}). Following~\cite{powers_effects_2018, mertan_modular_2023}, we use the minimum performance among all morphologies as the fitness for the evolutionary algorithm to help avoid specialization to a subset of morphologies while ignoring others. 

\begin{figure}
    \centering
    \begin{subfigure}[t]{\linewidth}
         \centering
         \includegraphics[width=\linewidth]{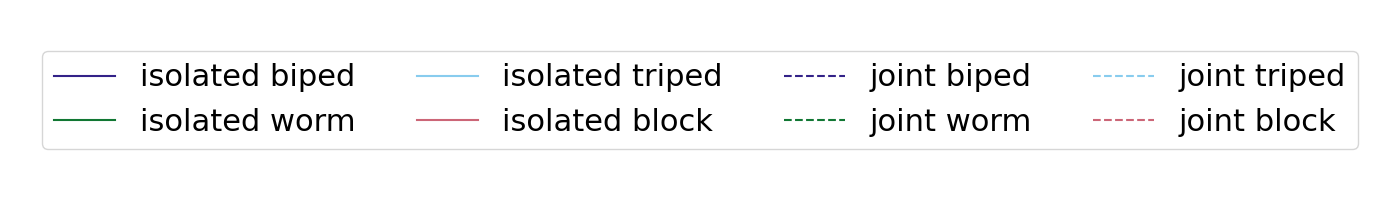}
    \end{subfigure}\\
    \begin{subfigure}[t]{\linewidth}
         \vspace{-19pt}
         \includegraphics[width=\linewidth]{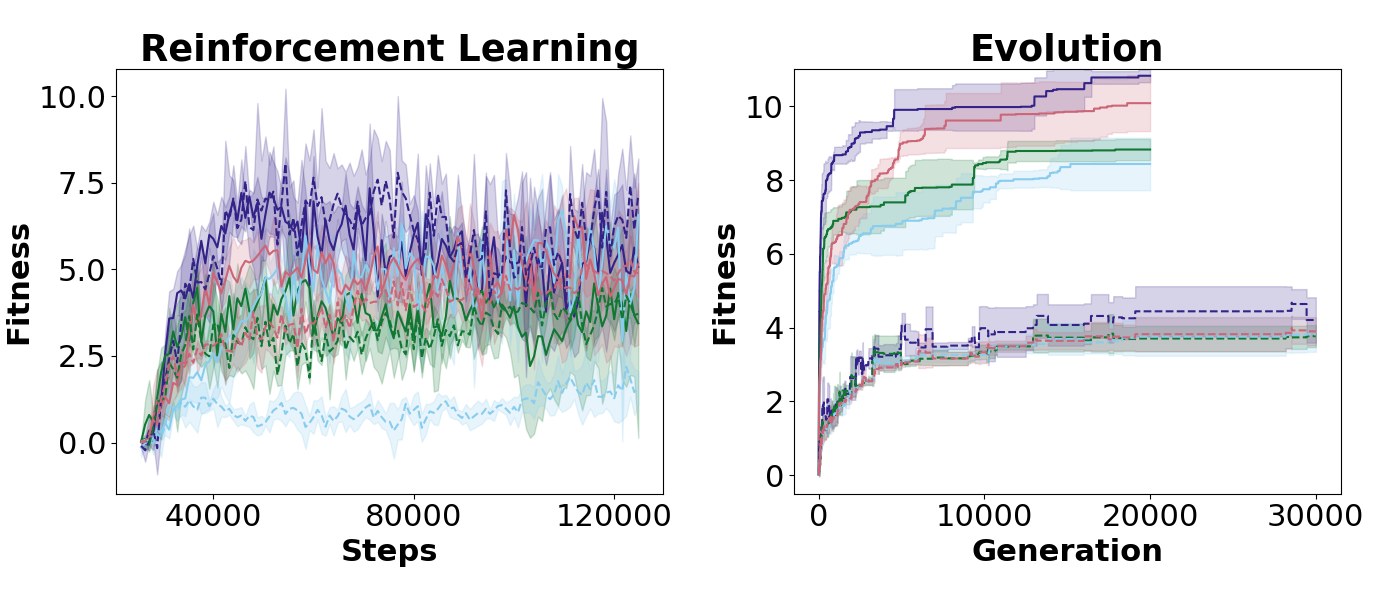}
    \end{subfigure}
    \caption{Training trajectories of isolated training on each morphology individually (solid lines) vs. the performance of each morphology during joint training on all (dashed lines). Lines show the mean values and the shaded areas show the standard errors, calculated over 3 repetitions. Reinforcement Learning (left) can find solutions that work well for multiple morphologies but ignore others. Evolutionary algorithms (right) find solutions that perform similarly on all morphologies, but they exhibit sub-optimal performance. }
    \label{fig:naive_app}
\end{figure}

Fig.~\ref{fig:naive_app} demonstrates the joint training trajectories with both algorithms, as well as isolated training for each morphology individually. Both algorithms struggle to optimize a multi-morphology controller during joint training. While the evolutionary algorithm converges to a sub-optimal but similar-performing solution for all morphologies, reinforcement learning finds solutions that work better for some morphologies, ignoring others. Yet both algorithms fail to find controllers that match the performance of single-morphology controllers each trained in isolation. 

To overcome the challenging training of multi-morphology controllers, here we propose the use of response-based, offline, multi-teacher knowledge distillation~\cite{buciluǎ2006model,hinton2015distilling}, also known as policy distillation in reinforcement learning~\cite{rusu2015policy,parisotto2015actor}, for the training of a single general controller capable of controlling multiple robots with different morphologies. We evolve single-morphology controllers for our experimented morphologies in Fig.~\ref{fig:fixed_ms} and use their experiences to create a dataset of $(observation,\allowbreak action)$ pairs. This dataset is then used for supervised learning to train a single multi-morphology controller to minimize the error in recreating the correct action for a given observation regardless of which robot the action is sampled from. Once we have the dataset, the training of the multi-morphology controller can be done in an offline supervised fashion. In our experiments, we model the controllers by neural networks and use the gradient descent algorithm Adam~\cite{kingma2014adam} to train them. 

To demonstrate the effectiveness of this approach, we replay the best controllers found by the evolutionary algorithm during isolated training of each morphology at each repetition, $C_m^r, m \in \{Biped, Worm, Triped, Block\}, r \in \{1, 2, 3\}$, 100 times and collect the $(observation, action)$ pairs into 81 ($3^4$, all possible combinations of champions for each morphology) datasets $D^i, i \in [1..81]$. Using these datasets, we distill 81 multi-morphology controllers, all with the same controller model as the teacher controllers -- Global FC,
by training the controllers on the datasets for 100,000 steps with mini-batches of size 128 and learning rate of 0.001, in a supervised manner where the loss is the mean squared error between the estimated and ground truth actions. All hyperparameters are chosen heuristically.

\begin{figure}
    \centering
    \includegraphics[width=0.8\linewidth]{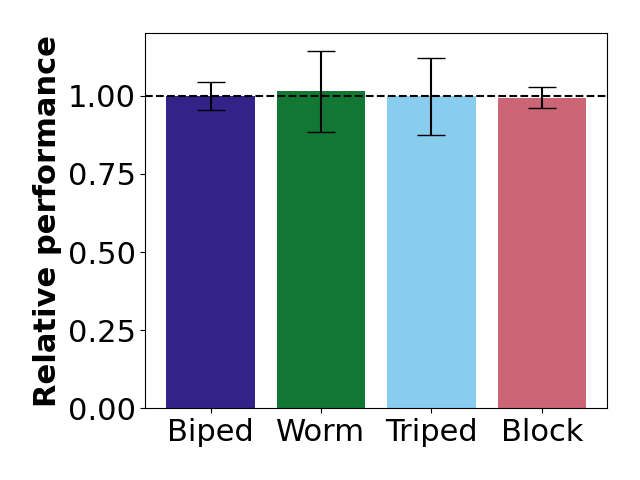}
    \caption{Performance of the multi-morphology controller relative to teacher single-morphology controllers for each experimented morphology ($mean \pm SE$). Here, and in all figures, the dotted line marks equal performance. Knowledge distillation can successfully train a single controller to control multiple morphologies as well as controllers specifically optimized for individual morphologies.}
    \label{fig:kd-multiVSsingle}
\end{figure}

Fig.~\ref{fig:kd-multiVSsingle} shows the performance of the distilled multi-morphology controllers relative to the corresponding single-morphology controllers, averaging over 81 cases. Knowledge distillation from single-morphology controllers results in multi-morphology controllers that achieve near-perfect teacher-level performance. Moreover, we see that in some cases the distilled multi-morphology controller outperforms the teacher single-morphology controller, as indicated by the error bars. These results demonstrate the effectiveness of knowledge distillation for the training of compatible controllers, without resorting to complex architectures or training schemes. 

\section{Quality Diversity for Domain Exploration} \label{sect:qd}
In the previous section, we experiment with heuristically chosen fixed morphologies and show that knowledge distillation can be utilized successfully to train a single controller capable of controlling multiple robot morphologies, without the need for any specialized architecture. Now the question is, how should we choose which morphologies to use for the training of the teacher single-morphology controllers, given a domain -- an environment and a task?

As opposed to using heuristically chosen morphologies~\cite{huang_one_2020,kurin_my_2021,nadizar2023fully}, we propose to utilize the QD algorithms~\cite{pugh_confronting_2015,pugh_quality_2016} to evolve distinct high-performing solutions representing trade-offs in a feature space, exploring the solution space for a given domain. Defining the feature descriptors based on the morphology of the robot, we can utilize QD algorithms to evolve different robot-controller pairs with varying morphologies, optimized for a particular environment and task. We prefer the QD algorithm over alternatives, as it allows finer control over the diversity of morphologies by explicitly describing feature descriptors and bins, and results in greater diversity~\cite{nordmoen2021map}. Knowledge distillation can then be applied to create a single controller capable of controlling a variety of different robots, exhibiting behavior optimized for the domain.

\begin{figure}
    \centering
    \includegraphics[width=0.8\linewidth]{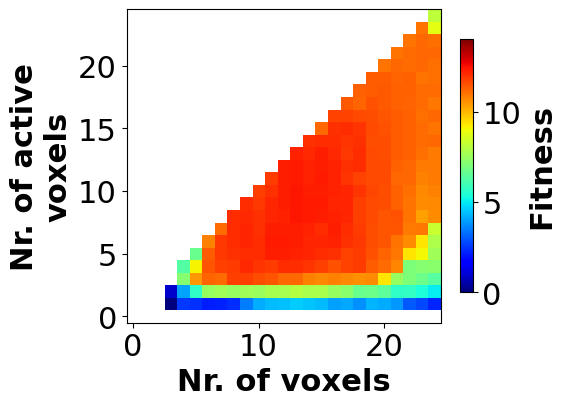}
    \caption{An example map produced by the MAP-Elites algorithm. Each cell corresponds to a robot-controller pair with its fitness shown as the color of the cell. The X-axis differentiates bins in the map by the number of total voxels present in the robot, while the y-axis stratifies robots by their number of active voxels. MAP-Elites successfully evolves a variety of high-performing robots.}
    \label{fig:qd-map}
\end{figure}

In particular, we experiment with the MAP-Elites algorithm~\cite{mouret_illuminating_2015}, similar to~\cite{nordmoen2021map}, where the feature descriptors for robots are the number of existing voxels and the number of active voxels. Fig.~\ref{fig:qd-map} shows the map produced by the MAP-Elites algorithm after 20,000 generations of evolution. We see that
the MAP-Elites algorithm explores the morphology space for the given task and evolves a map with 296 wide-ranging unique morphologies that can locomote as effectively as the ones in similar works~\cite{mertan_modular_2023,mertan2024investigating}.

Having produced this map for a given task or domain, we start experimenting with knowledge distillation to create a multi-morphology controller. We experiment with two heuristically chosen criteria for the selection of teacher robots: fitness and morphology. 

\begin{figure}
    \centering
    \begin{subfigure}[t]{0.55\linewidth}
         \centering
         \includegraphics[width=\linewidth]{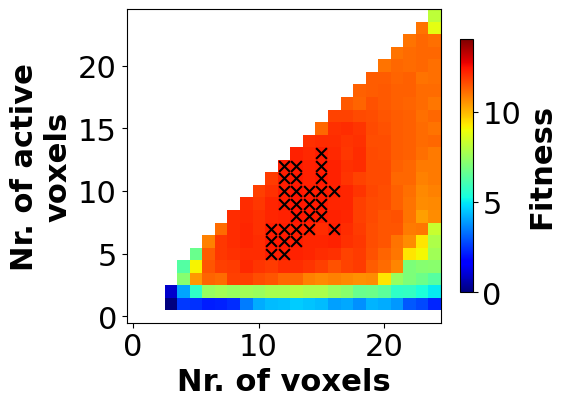}
    \end{subfigure}%
    \begin{subfigure}[t]{0.35\linewidth}
         \centering
         \includegraphics[width=\linewidth]{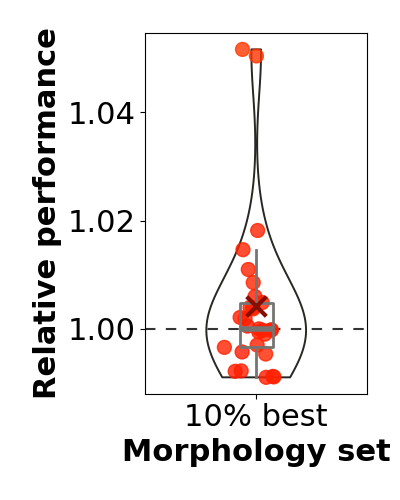}
    \end{subfigure}
    \caption{(left) Top 10\% of individuals (29 in total) used as teachers for distilling a multi-morphology controller, marked with an x. (right) Performance of the distilled multi-morphology controller on each trained morphology compared to their original teacher controllers across 10 runs with noise. Each data point is plotted and its color represents the fitness of the original controller. The mean point is labeled with an x. The distilled controller achieves almost perfect performance, matching the performances of teacher single-morphology controllers.}
    \label{fig:10-percent}
\end{figure}

First, we experiment with selecting different robots as teachers based on their fitness values. We order the solutions by their fitness values and distill a multi-morphology controller for the top $10\%$ of solutions, which are shown in Fig.~\ref{fig:10-percent} (left). The distilled multi-morphology controller successfully controls 29 slightly different morphologies representing different trade-offs between the number of voxels and the number of active voxels, and its performance is statistically indistinguishable  (at $p=0.05$-level) from the single-morphology controllers as shown in Fig.~\ref{fig:10-percent} (right).

\begin{figure}
    \centering
    \begin{subfigure}[t]{0.55\linewidth}
         \centering
         \includegraphics[width=\linewidth]{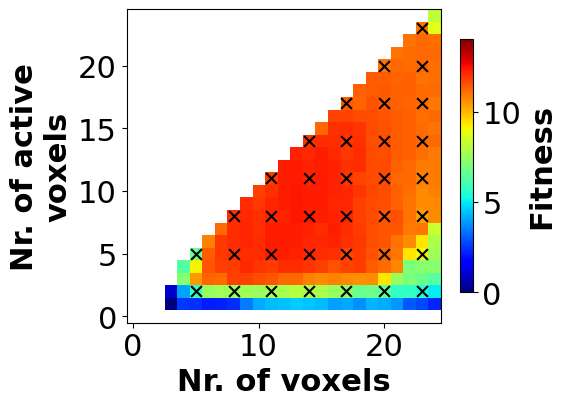}
    \end{subfigure}%
    \begin{subfigure}[t]{0.35\linewidth}
         \centering
         \includegraphics[width=\linewidth]{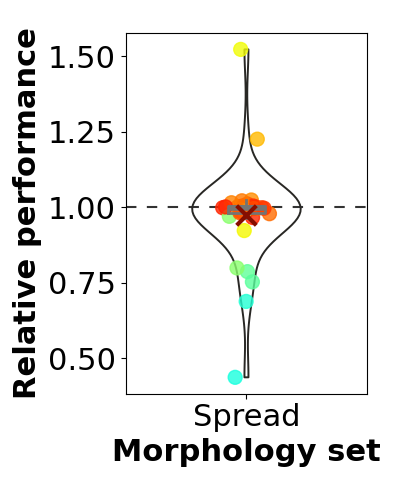}
    \end{subfigure}
    \caption{(left) 35 individuals maximally spread across the feature space, marked with an x. (right) Performance of the distilled multi-morphology controller on each trained morphology compared to their original teacher controllers across 10 runs with noise. The distilled controller matches the performance of teacher controllers in most cases. The cases where there is a performance drop occur where the teacher controller does not perform well on the morphology.\protect\footnotemark}
    \label{fig:qd-spread}
\end{figure}

To test the whether distillation process can result in controllers capable of controlling maximally different morphologies, we choose 35 individuals from the map that are spread across the feature space, as shown in Fig.~\ref{fig:qd-spread} (left). The performance of the distilled controller compared to teacher controllers on each experimented morphology can be seen in Fig.~\ref{fig:qd-spread} (right). While the controller is capable of matching the performance of the teacher controllers in 28 out of 35 morphologies ($p>.32$), it performs worse than the teachers in 5 cases and exceeds the teacher performance in 2 cases. Surprisingly, the cases where the distilled controller fails to achieve teacher-level performance occur when the teachers' performances are low (color represents the fitness of the original controller). 

\footnotetext{See the distilled controller \href{https://github.com/mertan-a/towards-multi-morphology-controllers?tab=readme-ov-file\#watch-the-behaviors-of-the-robots}{in action.}}

Overall, our experimentation shows that QD algorithms can be utilized to discover effective morphologies, as well as controllers optimized to control them. Subsequently, the knowledge that resides in the map can be distilled into a single controller, resulting in a controller capable of controlling a diverse set of morphologies.

\section{Investigating Abilities of Distilled Controllers} \label{sect:invest}
\subsection{Model Independence}\label{subsect:model_independent}
So far we have experimented with the same controller model for the distilled controller as the teacher controller -- Global FC.
However, once we collect the dataset of $\{observation,\allowbreak action\}$ pairs, we can train any controller architecture to do the mapping. It allows us to use deeper or more complex architectures only for the distillation part of the process where we use supervised learning while using simpler architectures that can be effectively trained with evolution during the QD process. Moreover, the distilled controller is not limited to the same control paradigm as the teachers either. For instance, the dataset can be pre-processed to turn global observations into local observations to distill a shared decentralized/modular controller (similar to the ones used in~\cite{pigozzi_evolving_2022, medvet_evolution_2020, medvet_biodiversity_2021, huang_one_2020,mertan_modular_2023,mertan2024investigating}) while the teachers are centralized/global controllers (similar to the ones used in~\cite{medvet_evolution_2020, ferigo_evolving_2021, mertan_modular_2023, talamini_evolutionary_2019, talamini_criticality-driven_2021,medvet_impact_2022}). 

\begin{figure}
    \centering
    \begin{subfigure}[t]{0.33\linewidth}
         \centering
         \includegraphics[width=\linewidth]{figures/qd_figures/distill_qd_spread_experiment.png}
         \caption{Global FC}
    \end{subfigure}%
    \begin{subfigure}[t]{0.33\linewidth}
         \centering
         \includegraphics[width=\linewidth]{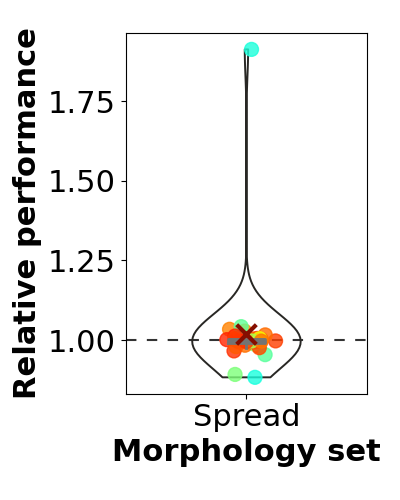}
         \caption{Global Tx}
    \end{subfigure}%
    \begin{subfigure}[t]{0.33\linewidth}
         \centering
         \includegraphics[width=\linewidth]{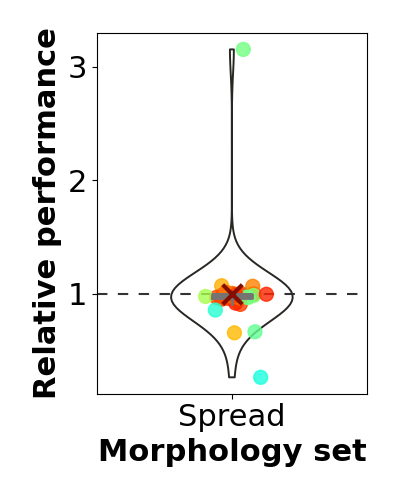}
         \caption{Modular FC}
    \end{subfigure}
    \caption{The same dataset acquired from individuals that are spread across the feature space distilled into three different controllers. (left) The same controller as the teacher controller achieves the same level of performance as the teachers for most of the cases. (middle) We distill into a more capable transformer controller and show that it is capable of achieving teacher-level performance in all cases, even surpassing teacher performance in one case. (right) We process the dataset to change the global observations to local observations and distill a modular controller that achieves a similar multi-morphology performance as the global controller. Please note the varied y-axis scales across subfigures.}
    \label{fig:qd-controllers}
\end{figure}

To display controller model independence, we use the 35 individuals that are maximally spread to the feature space, shown in Fig.~\ref{fig:qd-spread} (left), as teachers and distill a transformer-based global controller ("Global Tx") as well as a fully-connected neural network based modular controller ("Modular FC). All controllers achieve near-teacher-level performance on almost all morphologies, as shown in Fig.~\ref{fig:qd-controllers}. Moreover, their performances show variation, indicating that the type of controller affects the distillation performance. This creates an opportunity for incorporating existing work that designs compatible multi-morphology architectures~\cite{kurin_my_2021, gupta2022metamorph, wang2018nervenet,huang_one_2020, christensen_distributed_2013} into our approach to improve multi-morphology performance further.


\subsection{Scaling}\label{subsect:scale}

To test how many different teacher individuals can be distilled into a single multi-morphology controller, we experiment with using increasing numbers of individuals from the map as teachers. Specifically, we experiment with using the top $10, 40, 75, 100\%$ of individuals in the order of their fitness as teachers, which results in 29, 118, 222, and 296 teachers, respectively.

\begin{figure}
    \centering
    \includegraphics[width=0.8\linewidth]{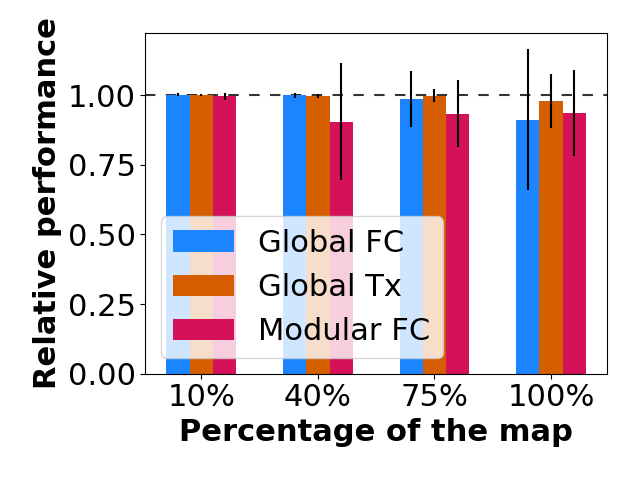}
    \caption{Performance of the distilled controllers with an increasing number of teachers with different morphologies. While different distilled controllers scale differently, they all achieve similar performances compared to teachers.}
    \label{fig:scaling}
\end{figure}

When we distill into all of the introduced multi-morphology controllers (Global FC, Global Tx, Modular FC), we see that all of the distilled controllers achieve performances closer to teachers, as shown in Fig.~\ref{fig:scaling}. As the number of teachers increases, the performance for the distilled controllers tends to slightly decrease and the error bars grow. However, we see that they show different scaling behaviors, demonstrating the possibility that a bigger or more complex controller architecture can achieve better performance. While we experiment with heuristically chosen architectures as a proof of concept, one can treat the architecture as a hyperparameter and optimize it for a particular domain.

\subsection{Generalization to Unseen Morphologies}\label{subsect:generalize}

We have shown above that (1) we can evolve a diverse set of morphologies with their respective controllers for the locomotion task optimized for different trade-offs in the feature space, and (2) these individuals can be used as teachers to distill into a single multi-morphology controller. The distilled controller is capable of controlling a diverse set of morphologies with near teacher-level performance and scaling up to the full map that covers 296 distinct morphologies. Here, we examine whether distilled controllers can generalize to unseen morphologies. 

\begin{figure}
    \centering
    \begin{subfigure}[t]{0.4\linewidth}
         \centering
         \raisebox{10.1pt}{\includegraphics[width=\linewidth]{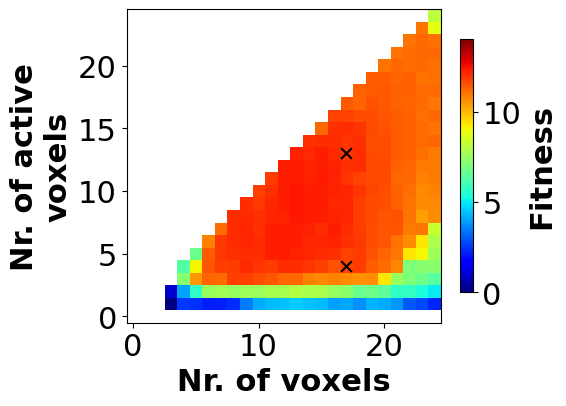}}
    \end{subfigure}
    \begin{subfigure}[t]{0.6\linewidth}
         \centering
         \includegraphics[width=\linewidth]{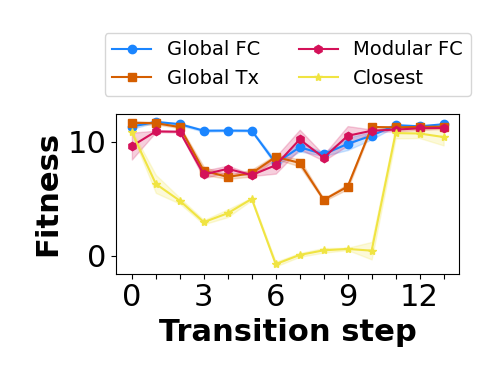}
    \end{subfigure}
    \caption{Using the pair of individuals marked on the map (left), we create a list of unseen morphologies in between the two morphologies of the selected individuals and measure the performance of the distilled controllers (right). As a baseline, we also measure the performance of the controller with the closest morphology from the map. Solid lines show the mean values of 10 fitness evaluations and shady regions show the standard errors. Distilled controllers generalize well to unseen morphologies, never performing worse than the closest controller from the map and outperforming them in most cases (transition steps 1 to 10).}
    \label{fig:unseen}
\end{figure}

To test the generalization of the distilled controller to unseen morphologies, we choose two random individuals from the map, shown in Fig.~\ref{fig:unseen} (left), 
and create a list of morphologies that transition from the morphology of one individual to the other, by changing one voxel at a time. We discard any morphology if it is not connected or if it exists in the map. This list of morphologies contains a number of unseen morphologies between the chosen individuals. We also note that this process can create unseen morphologies that are missing a number of voxels from seen morphologies and can be considered amputation scenarios. Therefore testing controllers on these morphologies also provides a way for us to test the damage recovery abilities of the controllers. As opposed to methods like \cite{cully_robots_2015, kriegman_automated_2019, bongard_resilient_2006} where the system spends time to figure out how to recover from damage, we would be measuring instant (i.e. zero-shot) damage recovery through the robustness of the controller.

To this end, we test distilled controllers (trained on the full map) on this set, as well as the single controller from the map was trained on the body plan closest to new morphologies. Fig.~\ref{fig:unseen} (right) shows a typical example of this process. While the performances of the closest controllers from the map drop as the morphology changes, distilled controllers perform better than the baseline. Moreover, their performance differs based on the controller, demonstrating the possibility that a specialized controller architecture or control strategy can be even more effective in its generalization capabilities.

\begin{figure}
    \centering
    \includegraphics[width=\linewidth]{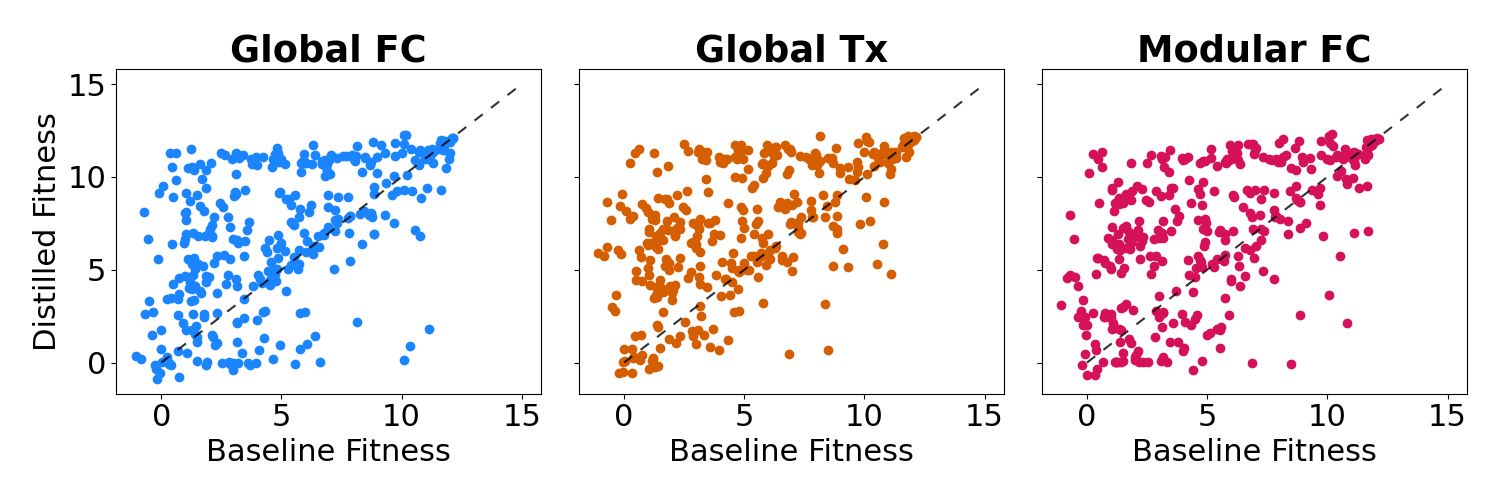}
    \caption{Performances of the distilled controllers plotted against the baseline controllers on 357 unseen morphologies. 
    Each data point is plotted. The distilled controllers outperform the baseline in most cases (all $p<.001$), showing the emergent ability to instantly adapt to unseen morphologies.}
    \label{fig:unseen-repeat}
\end{figure}

We repeat this process with 30 randomly chosen individual pairs and report the performances of distilled and baseline controllers on each unseen morphology (357 in total) in Fig.~\ref{fig:unseen-repeat}. Points above the dotted line are the cases where the distilled controllers outperformed the controller of the most similar morphology from the map (the number of such cases is 250 for the Global FC, 267 for Global Tx, and 246 for Modular FC). Overall, the distilled controllers perform approximately 1.5 times better than the closest controller on average, outperforming the baseline (all $p<.001$ with the alternative hypothesis that the sample distribution mean is greater than 1) and showing 
ability to adapt to unseen morphologies 
instantly.

\subsection{Rapid Finetuning}\label{subsect:prior}

The above results show impressive instant generalization to new morphologies interpolated within the map. But what about adaptation to these morphologies via further controller optimization? We are interested in testing how good of a prior the distilled controllers are for further specialization on unseen morphologies or on different tasks. Ideally, we expect the distilled controller to perform better as a starting point for further specialization, allowing us to have foundational models that enable rapid adaptation to downstream tasks, whether they are unseen morphologies or different objectives (akin to initializing an image classifier with ResNets pre-trained on ImageNet). This can help mitigate the cost of training for many robotics applications and accelerate the development in the field.

To see how good of a prior the distilled controller is for further specialization on unseen morphologies and tasks, we run 300 generations of AFPO~\cite{schmidt_age-fitness_2010} where all individuals in the starting population are initialized with the Global FC multi-morphology controller distilled from the full map. For each unseen morphology that we sampled in the previous experiment (357 in total), we fix the morphology and finetune the distilled controller on that morphology performing the original Walker-v0 task.  To assess wider generalization, in a second condition, we also take that same morphology and controller originally learned on the Walker-v0 task and finetune it on the BridgeWalker-v0 environment (a slightly different locomotion task). As a baseline, for each unseen morphology, we find the controller of the robot with the most similar morphology from the map and initialize the evolutionary run with that controller instead of the distilled controller. We note that both the distilled controller and the baseline controllers have the same architecture.

\begin{figure}
    \centering
    \begin{subfigure}[t]{0.5\linewidth}
         \centering
         \includegraphics[width=\linewidth]{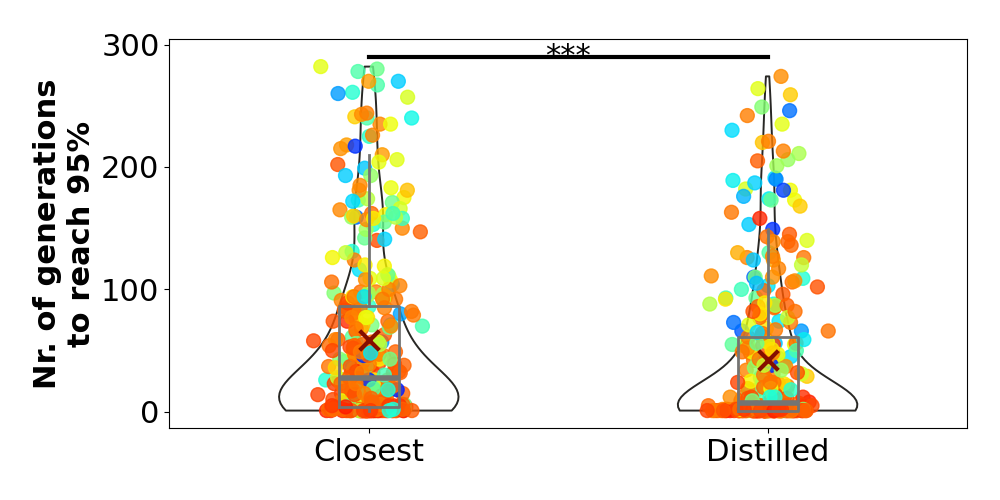}
         \caption{Walker-v0}
    \end{subfigure}%
    \begin{subfigure}[t]{0.5\linewidth}
         \centering
         \includegraphics[width=\linewidth]{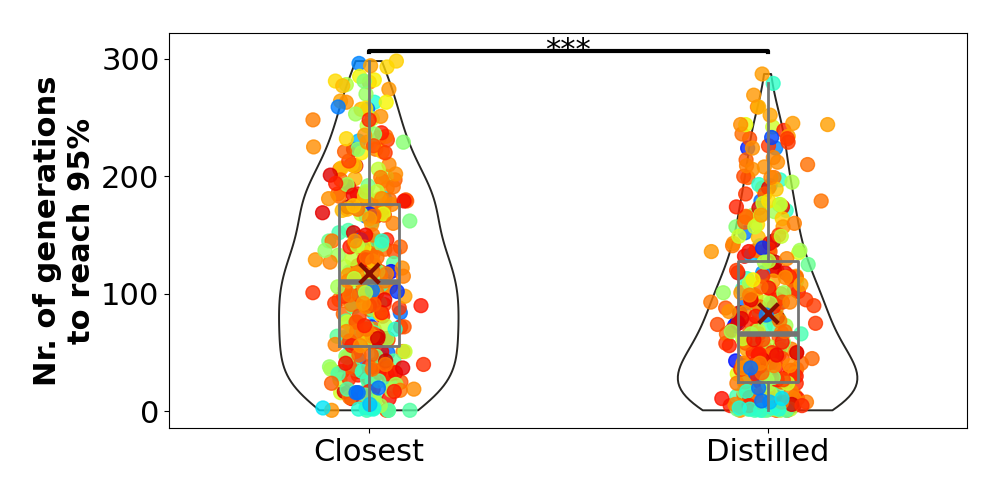}
         \caption{BridgeWalker-v0}
    \end{subfigure}
    \caption{Number of generations to achieve 95\% of the end performance for the distilled controller and the baseline controller, lower is better. The distilled controller converges slightly faster compared to the baseline controllers. This trend also held in the time to reach 90\% or 99\% of the end performance (all $p<.01$). 
    }
    \label{fig:prior_speed}
\end{figure}

While the performances of the run champions do not show any statistically significant differences between the baseline and the distilled controller in both environments, we see that runs with the distilled controller achieve 90\%, 95\%, and 99\% of their respective final performances faster compared to runs with the baseline controllers (Fig.~\ref{fig:prior_speed}). These results demonstrate that the distilled controller provides a better starting point for further specialization, enabling rapid finetuning to downstream tasks.  


\section{Discussion} \label{sect:disc}
The first part of our pipeline consists of QD algorithms for the discovery of high-performing diverse morphologies and their controllers. While this solves the issue of finding good morphologies to train a distilled model on, which has not been addressed in previous work, it also increases the computational cost of our pipeline greatly, since we end up optimizing a controller for each morphology in the map. While we believe our method has an advantage in this regard since we can use simple controllers in the QD phase and then distill their knowledge into any complex model, we leave the comparison of our method to methods such as~\cite{huang_one_2020,gupta2022metamorph,nadizar2023fully} in terms of number of simulation steps for future work. 

In our QD experiments, we observed that the number of migrations (i.e. solutions that move from one cell of the MAP-Elites map to another via a mutation to their morphology) is very limited. This is in line with the existing literature in brain-body co-optimization that demonstrates mutation to morphologies is often detrimental to the performance and results in ineffective search over the morphology space~\cite{cheney_difficulty_2016,cheney_scalable_2018,mertan_modular_2023,mertan2024investigating}. The inability of solutions to migrate to new morphologies is concerning as goal-switching and creating stepping stones are core principles that make QD algorithms, and especially MAP-Elites, effective and important. We believe that having a relatively high-resolution map was helpful in our case, as it reduces the difference between morphologies in neighboring cells.  However, we are doubtful how well QD algorithms will scale to more complex morphology spaces where having a higher-resolution map would be infeasible. In future work, we are going to investigate this phenomenon and how QD algorithms behave on the problem of brain-body co-optimization. 

The second part of our pipeline is knowledge distillation as an alternative for the training of multi-morphology controllers (given that there exist teacher controllers). Our investigation shows that knowledge distillation can train simple controllers for multiple morphologies that would be untrainable from scratch without the capacity of a much larger network. Crucially, the distilled controllers show emergent properties. The most important of them is the generalization to unseen morphologies in a zero-shot manner. The generalization ability allows distilled controllers to be robust to perturbations to the morphology of the robot. In this sense, we consider our work as the successor of the "Resilient Machines"~\cite{bongard_resilient_2006} and "Robots that can adapt like animals"~\cite{cully_robots_2015}. In the former, the adaptation to morphological changes happens through continuous self-modeling that happens in an evolutionary time scale, and in the latter, the adaptation occurs in a faster time scale through intelligent search over a behavior repertoire. In our case, the distilled controller is already capable of controlling multiple morphologies (including seen and unseen morphologies) and can adapt instantly. In the case of a failure, one can work backward in the methods to find or optimize a controller that can recover from the damage.

Moreover, we are interested in utilizing distilled controllers to enhance the search over the morphology space. The literature on brain-body co-optimization points out the ineffective search over the morphology space due to fragile co-adaptation of body and brain as a major challenge for brain-body co-optimization~\cite{cheney_difficulty_2016,mertan2024investigating, pfeifer2006body}. Recently, the investigation of~\cite{mertan2024investigating} indicates that not being able to estimate the maximum performance of morphology without fully optimizing a controller may be a critical part of what makes the search ineffective. The distilled multi-morphology controller that generalizes to unseen morphologies can alleviate this issue and enable a better search over the morphology space.

In a similar vein, being able to control a multitude of modular morphologies unlocks a potential for adaptation and functionality through morphological changes. A distilled general controller can enable ideas such as damage recovery through shape-shifting, and reconfiguration to perform different functions similar to the ones in~\cite{kriegman_automated_2019,zykov2005self,zykov_evolved_2007}. We would like to investigate the ways of exploiting general controllers in these ways in future work.

In future work, we would like to examine the generalization ability of the distilled controllers when we use more complex compatible controllers~\cite{gupta2022metamorph,huang_one_2020,wang2018nervenet,kurin_my_2021}. Moreover, we used heuristically chosen parameters for the distillation process where we trained the distilled controller for a fixed number of steps. We would like to investigate the use of a held-out validation set of morphologies to maximize the generalization abilities via for early stopping or meta-learning of controllers~\cite{finn_model-agnostic_2017, sun2019meta}. We also experimented with heuristically chosen ways of selecting teachers and assumed that a multi-morphology controller distilled from the full map would be the best for their emergent properties such as generalization. However, it is not clear how many teacher robots we need to effectively train a distilled general controller, or which set of pre-trained robots make the best teachers. Future work should investigate how the selection of teachers affects the distilled controllers' performance and how they should be selected. Lastly, we used the Evogym simulator~\cite{bhatia2021evolution}, which is a 2D voxel-based soft robot simulator. The applicability of our method in different types of robots/domains, in 3D, and in real robot scenarios should be investigated in future work to show the generality of our approach.

\section{Conclusion} \label{sect:conc}
We present a pipeline for creating compatible controllers that can control multiple morphologies. Given an environment and a task, we explore with Quality Diversity algorithms~\cite{pugh_confronting_2015,pugh_quality_2016} to find morphologies representing different trade-offs and individual controllers optimized to control these morphologies. After the exploration phase, we exploit these specialized single-morphology controllers as teacher controllers and distill their behaviors into a single controller that can control multiple morphologies via supervised knowledge distillation. We find that distilled controllers generalize well to unseen morphologies, and increase adaptation efficiency as a prior for further finetuning of the behavior.  We hope that this approach will contribute to future progress in multi-robot general controllers and brain-body co-optimization.  

\begin{acks}
This material is based upon work supported by the National Science Foundation under Grant No. 2008413, 2239691, 2218063.
Computations were performed on the Vermont Advanced Computing Core supported in part by NSF Award No. OAC-1827314.
\end{acks}

\bibliographystyle{ACM-Reference-Format}
\bibliography{sample-base}


\end{document}